\documentclass[letterpaper, 10 pt, conference]{ieeeconf}  

\IEEEoverridecommandlockouts                              

\usepackage{graphics} 
\usepackage{epsfig} 
\usepackage{times} 
\usepackage{amsmath} 
\usepackage{amssymb}  
\usepackage[makeroom]{cancel}

\usepackage{geometry}
\usepackage{shortcuts}

\makeatletter
\let\NAT@parse\undefined
\makeatother
\usepackage[urlcolor=blue]{hyperref} 

\title{\LARGE \bf
Synthesizing Robust Walking Gaits via Discrete-Time Barrier Functions with Application to Multi-Contact Exoskeleton Locomotion} 
\author{Maegan Tucker$^{1}$, Kejun Li$^{2}$, and Aaron D. Ames$^{1,3}$
\thanks{This research was supported by Wandercraft and the Zeitlin Family Fund. Research involving human subjects was conducted under IRB No. 21-0693}%
\thanks{$^{1}$ Maegan Tucker is with the School of Electrical and Computer Engineering and the Woodruff School of Mechanical Engineering, Georgia Institute of Technology,
Atlanta, GA 30308.}%
\thanks{$^{2}$ Kejun Li is with the Department of Computation and Neural Systems, California Institute of Technology,
Pasadena, CA 91125.}
\thanks{$^{3}$ Aaron D. Ames is with the Department of Mechanical and Civil Engineering and the Department of Control and Dynamical Systems, California Institute of Technology,
Pasadena, CA 91125.}
\thanks{\texttt{mtucker@gatech.edu}}%
}

\begin{document}

\maketitle
\thispagestyle{empty}
\pagestyle{empty}

\begin{abstract}
Successfully achieving bipedal locomotion remains challenging due to real-world factors such as model uncertainty, random disturbances, and imperfect state estimation. In this work, we propose a novel metric for locomotive robustness -- the estimated size of the hybrid forward invariant set associated with the step-to-step dynamics. Here, the forward invariant set can be loosely interpreted as the region of attraction for the discrete-time dynamics. We illustrate the use of this metric towards synthesizing nominal walking gaits using a simulation-in-the-loop learning approach. Further, we leverage discrete-time barrier functions and a sampling-based approach to approximate sets that are maximally forward invariant. Lastly, we experimentally demonstrate that this approach results in successful locomotion for both flat-foot walking and multi-contact walking on the Atalante lower-body exoskeleton. 


\end{abstract}

\section{Introduction}



The field of robotic bipedal locomotion is receiving growing attention due to the widespread potential of humanoid robots. While this increase in attention is resulting in numerous demonstrations of experimentally stable robotic locomotion \cite{sreenath2011compliant, radford2015valkyrie, hubicki2016atrias, kuindersma2016optimization, bostondynamics2017, da2017supervised, xiong2020dynamic, paredes2022resolved, crowley2023optimizing}, it is still unclear how to systematically verify that the walking is robust to real-world disturbances, including those due to external perturbations, uncertain terrain, or uncertain models. 




In general, there are two main approaches for improving locomotive robustness. The first is via online planning either around the full system dynamics or a reduced-order model \cite{feng2016robust, garcia2021mpc}. These online planners are able to reject large disturbance by synthesizing motions in real time to return to certain desire terminal states. Since these planners are aimed at stabilizing the system to some prespecified terminal constraint, it is important to carefully consider how these references are constructed. 




This motivates the second method, directly improving the inherent robustness of nominal reference trajectories. Specifically, references that are more robust can improve the inherent robustness of the walking motion, thus decreasing the control effort required by online planners. Previous work towards improving robustness of reference trajectories leveraged methods from non-smooth analysis \cite{kong2023saltation} to generate limit cycles that are less sensitive to impact uncertainty \cite{zhu2022hybrid, tucker2023robust}. Similarly, previous work has synthesized robust limit cycles for uncertain impact events by minimizing the cost-to-go associated with early or late impacts \cite{dai2012optimizing}. While these approaches successfully yield robust limit cycles, they are not able to provide formal guarantees about the allowable disturbance bounds.  

\begin{figure}
    \centering
    \href{https://youtu.be/6aXsBKMxDH0}{\includegraphics[width=0.85\linewidth]{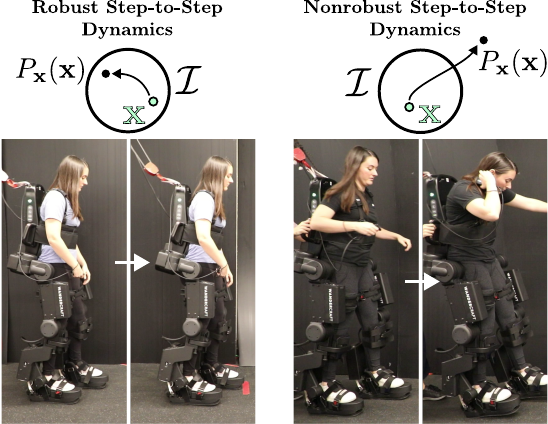}}
    \caption{The framework developed in this paper optimizes locomotive robustness using forward invariance, certified via discrete-time barrier functions, as a metric for robustness.}
    \label{fig: hero}
    \vspace{-4mm}
\end{figure}

There exist several methods to quantify locomotive robustness including computing the region of attraction \cite{manchester2011regions} and input-to-state stability \cite{kolathaya2018input, tucker2023input}. In contrast to the previous categories, these methods provide certificates of robustness, but are often not practical for behavior synthesis due to computational complexity.

In this paper, we take a step-to-step dynamics perspective \cite{bhounsule2020approximation, xiong20223}, transforming our system into a discrete-time system via the Poincar\'e return map. This perspective allows us to reason about the robustness of the full hybrid system by identifying forward-invariant sets centered around the fixed point of the Poincar\'e return map. The main motivation here is the observation that larger forward invariant sets correspond to larger regions of attraction. 
Towards this, we leverage discrete-time barrier functions and a sampling-based approach to approximate sets that are maximally forward invariant \cite{agrawal2017discrete, ahmadi2019safe, cosner2023robust}. 
The advantage of this approach is that it is general enough to be applied in combination with any control framework, while reflecting robustness and remaining computationally tractable for high-dimensional systems. This tractability is in part afforded by transforming the system into a reduced representation of the full-order model. 




Finally, we demonstrate how our proposed approach can be leveraged in a simulation-in-the-loop learning framework to systematically synthesize robust reference trajectories. The main purpose of this demonstration is to illustrate the effect of our proposed metric, without convoluting the robustness with the effect of online planners. However, a secondary purpose of this approach is to emphasize the role that reference trajectories have on the overall robustness of the walking motion. 





We experimentally deploy the framework for both flat-foot and multi-contact (foot-rolling) walking on the Atalante lower-body exoskeleton. This device is designed to restore locomotion to individuals with mobility impairments, including motor complete paraplegia. We also compare our results with a nominal approach which purely optimizes for heuristic stability.
Overall, the proposed method is shown to successfully synthesize robust nominal limit cycles in a principled fashion with computational tractability.

\section{Preliminaries on Locomotion}
\label{sec: prelim}
Walking naturally lends itself to be modeled as a hybrid system because of the presence of both continuous dynamics (during the swing phase) and discrete dynamics (at swing foot impacts). Further, the dynamics associated with periodic walking can be simplified to only consider a discrete-time hybrid system, evaluated either before or after impact events. The advantage of this representation is that the locomotive robustness can be evaluated by identifying forward-invariant sets of the discrete-time hybrid system. To motivate this viewpoint we will first introduce the notations.

\begin{figure}
    \centering
    \subfloat[Flat-foot behavior]{\includegraphics[width=0.35\linewidth]{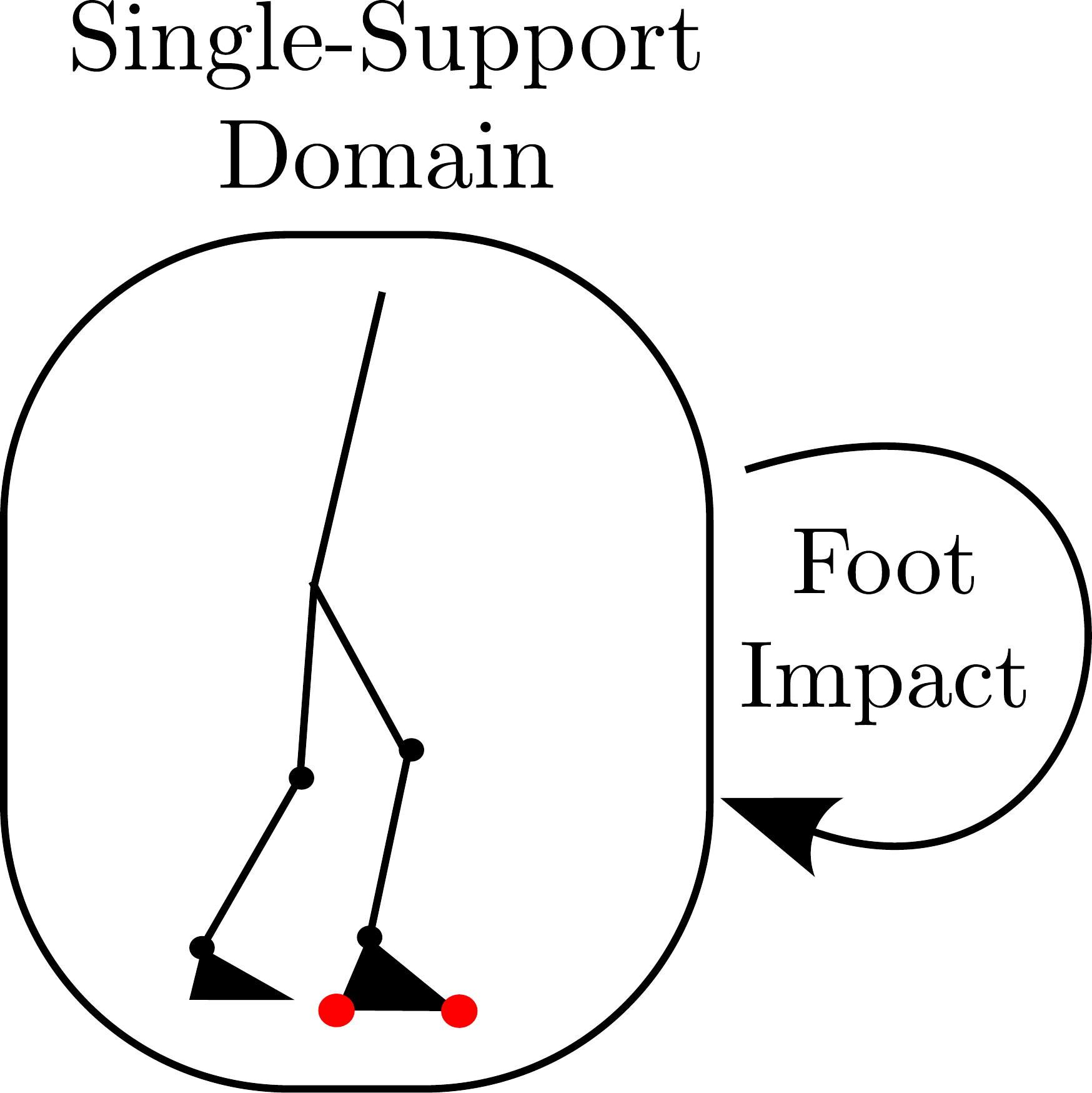}\label{fig: flatfoot-domain}}
    \hfill
    \subfloat[Multi-contact behavior]{\includegraphics[width=0.60\linewidth]{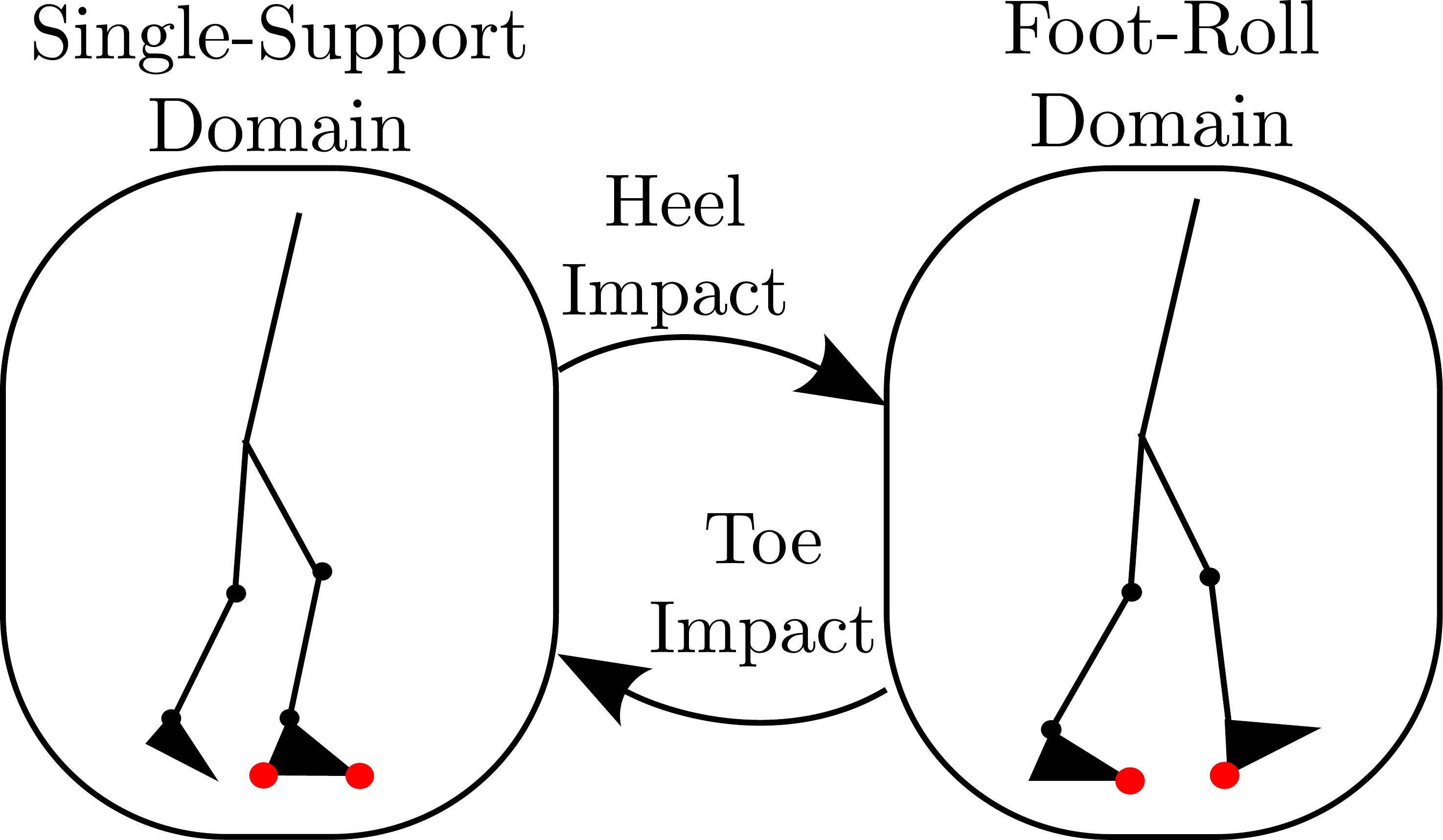}\label{fig: multi-domain}}
    \caption{Directed graphs describing the hybrid system domain structure for the a) flat-foot and b) multi-contact walking.}
    \label{fig: domaingraphs}
    \vspace{-4mm}
\end{figure}

\newsec{Hybrid Control Systems}
For simplicity, we will present hybrid control systems in the context of symmetric locomotion, with a single domain capturing the continuous-time dynamics of the swing phase and a single edge capturing the discrete-time dynamics of the swing-foot impact event. Note that this structure can be extended to capture more complicated domain structures \cite{reher2020algorithmic}. In our work, we will extend the domain structure to include a foot-rolling domain for multi-contact walking, as illustrated in Fig. \ref{fig: domaingraphs}.

Consider a system with states $x \in \X \subset \R^n$ and control input $u \in \mathcal{U} \subset \R^m$. For robotic systems, these states are typically defined as $x = (q_e^{\top},\dot{q}_e^{\top})^{\top}$, with $q_e := \{p_b^{\top}, \phi_{b}^{\top},q^{\top}\} \in \R^3 \times SO(3) \times \mathcal{Q}$ denoting the set of generalized coordinates of the system \cite{grizzle2014models}. 
The complete hybrid control system (for a system with a single domain and a single edge) can be modeled as:
\begin{numcases}{\mathcal{H}\mathcal{C}  = }
\dot{x} = f(x) + g(x) u & $x \in \D \setminus \S$, \label{eq: continuouscontrol}
\\
x^+ = \Delta(x^-) & $x^- \in \S$, \label{eq: discretecontrol}
\end{numcases}
where \eqref{eq: continuouscontrol} and \eqref{eq: discretecontrol} denote the continuous-time and discrete-time dynamics, respectively. Here, $\D \subset \X$ denotes the admissible domain on which the continuous-time dynamics evolve and $\S \subset \D$ denotes the \textit{guard}.
These sets are respectively defined as:
\begin{align}
    \D &= \{x\in\mathcal{X} \mid h(x) \ge 0\}, \\
    \S &= \{x\in \mathcal{X} \mid h(x) = 0,~\dot h(x) < 0\},
    \label{eq: zeroguard}
\end{align}
for some continuously differentiable function $h:\mathcal{X}\to\R$. 
Note that for bipedal robots, the function $h$ is typically defined to be the vertical height of the non-stance foot. It is also typically assumed that the ground height is known and constant. Lastly, it is assumed (as is typical) that all quantities in $\mathcal{H}\mathcal{C}$ are locally Lipshitz continuous, e.g., the impact map $\Delta$ is locally Lipschitz. This follows from the assumption of perfectly plastic impacts \cite{glocker1992dynamical}. 
For explicit forms of \eqref{eq: continuouscontrol} and \eqref{eq: discretecontrol} for robotic legged systems, we refer the reader to \cite{grizzle2014models}. 

\newsec{Synthesizing Nominal Limit Cycles via the HZD Method}
One approach towards achieving robotic locomotion is to synthesize nominal limit cycles, which when enforced, result in exponentially stable hybrid periodic orbits of the full system dynamics. One method for synthesizing these limit cycles is the Hybrid Zero Dynamics (HZD) method \cite{westervelt2003hybrid}. This method can also be extended to fully-actuated systems, with the extension termed the \textit{Partial} Hybrid Zero Dynamics (PHZD) method \cite{ames2012first}. 
In short, both methods synthesize nominal limit cycles using a trajectory optimization problem that enforces impact-invariance of the controllable outputs (termed virtual constraints),
defined as:
\begin{align}
    y(x,\beta) := y^a(x) - y^d(\tau(x),\beta), 
\end{align}
with $y^a: \X \to \R^m$ denoting the actual measured outputs and $y^d: [0,1] \times \R^{m\times (v+1)} \to \R^m$ denoting the desired outputs parameterized by a phasing variable $\tau: \X \to [0,1]$ and a collection of B\'ezier coefficients $\beta \in \R^{m \times (v+1)}$ with $v \in \N$ denoting the B\'ezier polynominal degree. By driving these outputs to zero exponentially, the hybrid control system is transformed into a closed-loop hybrid system:
\begin{numcases}
{\mathcal{H} = }
\dot{x} = f_{cl}(x) & $x \in \D \setminus \S$, \label{eq: closedloopcontinuous}
\\
x^+ = \Delta(x^-) & $x^- \in \S$. \label{eq: closedloopdiscrete}
\end{numcases}
Solutions of \eqref{eq: closedloopcontinuous} can be denoted using flow notation, $\varphi_t(x_0)$ evaluated at time $t \in \R^+$ with initial condition $x_0 \in \D$. This flow is periodic, with period $T \in \R^+$, if there exists a fixed point $x^* \in \S$ satisfying $\varphi_T(\Delta(x^*)) = x^*$. Thus, we arrive at the notation for the periodic orbit expressed using the flow:
\begin{align}
    \mathcal{O} := \{\varphi_t(\Delta(x^*)) \in \D \mid 0 \leq t \leq T_I \}, \label{eq: orbit}
\end{align}
with the time-to-impact function $T_I: \mathcal{X} \to \R$ defined as:
\begin{align}
    T_I(x) := \inf \{t \geq 0 \mid \varphi_t(\Delta(x^*)) \in \S \}.
\end{align}

\newsec{Discrete-Time System Representation}
Finally, the closed-loop hybrid system can be transformed into a discrete-time hybrid system via the Poincar\'e return map $P: \tilde{\S} \to \S$, with $\tilde{\S}$ denoting the region on which $T_I$ is well-defined. The Poincar\'e return map, denoted in flow notation as:
\begin{align}
    P(x^-) := \varphi_{T_I(x^-)}(\Delta(x^-)),
\end{align}
transforms the hybrid system into the discrete-time system:
\begin{align}
    x_{k+1} = P(x_k), ~k = 0,1,\dots.
\end{align}

The benefit of this discrete-time representation is the ability to analyze the stability of the overall system via stability of the step-to-step dynamics. In particular, as presented in Theorem 1 of \cite{morris2005restricted}, the periodic orbit $\O$ is exponentially stable if and only if for the fixed point $P(x^*) = x^*$, there exists $M>0$, $\alpha \in (0,1)$, and some $\delta > 0$ such that:
\begin{align}
    \forall ~ x \in B_{\delta}(x^*) \cap \widetilde{S} &\quad \implies \quad  \nonumber\\
 & \| P^i(x) - P(x^*)\| \leq M\alpha^i \|x - x^*\|, \nonumber
\end{align}
with $P^i(x)$ denoting the Poincar\'e map applied $i \in \N_{\geq 0}  = \{0,1,\dots,n,\dots\}$ times. 

However, stability of the periodic orbit alone does not elucidate the performance of the closed-loop system in the presence of impact uncertainty and disturbances. This motivates robustness measures, such as input-to-state stability (ISS) or input-to-state safety (ISSf). However, verifying these properties in practice remains computationally expensive for high-dimensional systems, illustrating the need for computationally-tractable measures of robustness.

\section{Robust Walking via Hybrid Forward Invariance}
\label{sec: robustness}

While the HZD method of gait generation has been successfully demonstrated in practice towards realizing stable locomotion on numerous robotic platforms, there exist relatively few methods towards characterizing the level of robustness of the generated gaits. Thus in practice, the gait generation is typically manually tuned until gaits are generated which produce robust locomotion on hardware. In this section, we will instead present a method for certifying forward invariance of the discrete-time system for a reduced representation of the full-order system.

\begin{figure}
    \centering
    \subfloat[Full system \label{1a}]{%
           \includegraphics[width=0.42\linewidth]{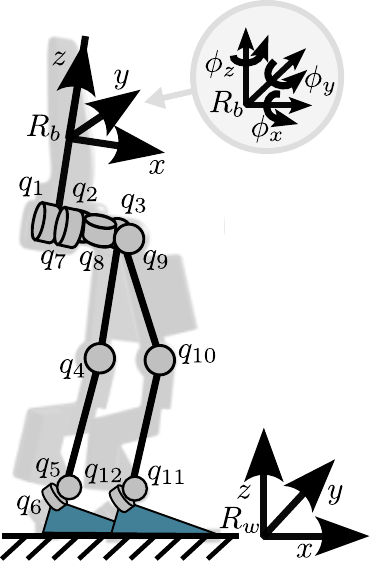}}
           \hspace{5mm}
      \subfloat[Reduced system \label{1b}]{%
            \includegraphics[width=0.35\linewidth]{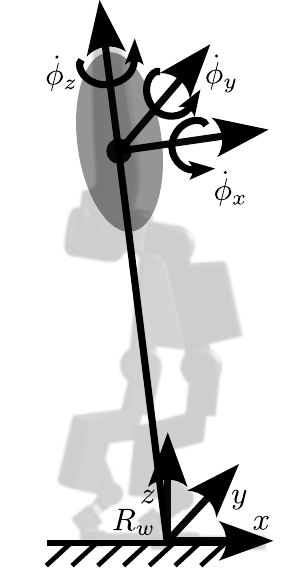}}
    \caption{Model representations. a) The full system model is denoted by the generalized coordinates $x = (q_e^{\top},\dot q_e^{\top})^{\top}$ with $q_e := (p_b^{\top},\phi_b^{\top},q^{\top})^{\top} \in \R^3 \times SO(3) \times \mathcal{Q}$. Here $p_b \in \R^3$ and $\phi_b$ respectively denote the euclidean position and orientation of the global base frame $R_b$ relative to the world frame $R_w$. b) Here, the reduced-order representation of the model is illustrated, defined as the angular velocities of the global frame relative to the world frame, i.e., $\x := (\dot \phi_x, \dot \phi_y, \dot \phi_z)^{\top}$.}
    \label{fig: reduce}
    \vspace{-2mm}
\end{figure}

\begin{figure*}
    \centering
    \includegraphics[width=\linewidth]{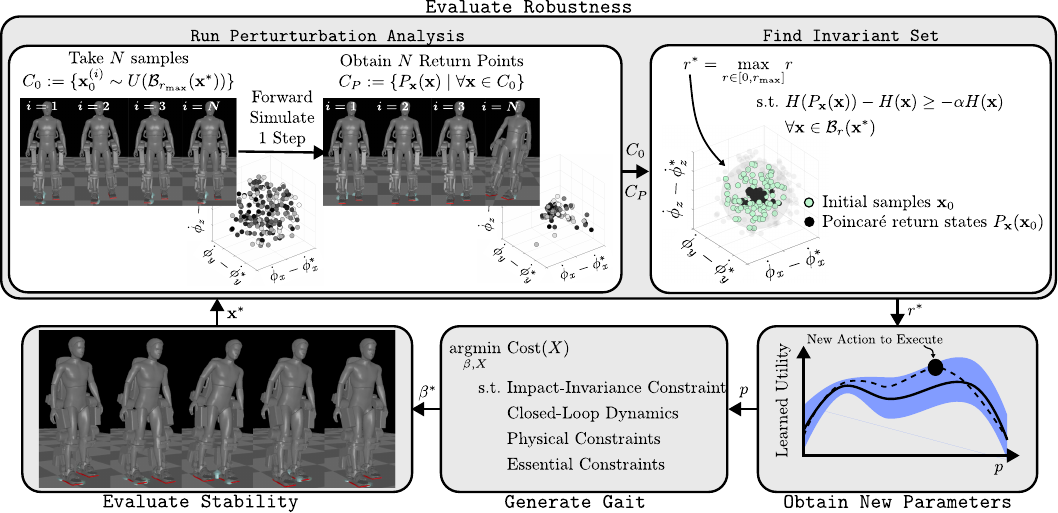}
    \caption{Diagram of sim-in-the-loop approach towards optimizing robustness.}
    \label{fig: diagram}
    \vspace{-2mm}
\end{figure*}

\newsec{Reduced System Representation}
While the following approach could be conducted for the full system state $x \in \X$, we propose the use of a lower-dimensional representation since certain coordinates often have a disproportional influence on overall system stability. For example, for bipedal robots, perturbations on actuated joints will influence the overall system much less than perturbations on the global coordinates. 
This viewpoint is similar to that of reduced-order models, whereby the essence of the full-order dynamics can be captured by a few key states. 

In general, the reduced state representation is denoted as $\x = \Phi(x) \in \mathbf{X}$, where $\mathbf{X}$ is a \emph{reduced-order manifold}, i.e., $\dim(\mathbf{X}) \leq \dim(\mathcal{D})$ consisting of the lower-dimension representation of interest. 


Assume that there exists a projection between our full system state and lower-dimension representation $\x = \Phi(x)$. For the Atalante lower-body exoskeleton, we restrict our attention to three specific global coordinates, denoted as the reduced system $\x := (\dot{\phi}_x,\dot{\phi}_y,\dot{\phi}_z)^{\top} \in \R^3$, representing the global angular velocity. This mapping is simply the labeling matrix $\x = \begin{bmatrix}
     \mathbf{0}_{3 \times 21} & \mathbf{I}_{3 \times 3} & \mathbf{0}_{3 \times 12}
    \end{bmatrix}x$. 
The motivation for selecting these coordinates is the observation that exoskeleton users often add perturbations to the system that can be captured by instantaneous angular velocity changes.
Note that in other applications, this reduced-order manifold could be the position of the center of the pressure in ZMP walking \cite{kajita2003biped}, centroidal dynamics \cite{dai2014whole, zhao2017robust}, or zero dynamics \cite{grizzle2008hzd}. We can represent the discrete-time system associated with this reduced model as the Poincar\'{e} map restricted to $\mathbf{X}$: 
\begin{align}
    \x_{k+1} = P_{\mathbf{X}}(\x_k)
    \label{eq: reduceddiscrete}
\end{align}
with $P_{\mathbf{X}} := \Phi(P(\iota(\x_k)))$ 
defined as the Poincar\'e return map for the reduced system for states on the ``reduced order'' guard $\mathbf{X} \cap \S$.   Here $\iota : \mathbf{X} \cap \S \to \S$ is a specific (non-unique) reconstruction of the full state, e.g. in the case when $\mathbf{X}$ are the zero dynamics, $\iota(\x)$ can be obtained from the outputs and their derivatives \cite{morris2005restricted}. 

\newsec{Discrete-Time Barrier Functions}
In this paper, we will utilize barrier functions to guarantee forward-invariance of our reduced hybrid system. Thus, we will first present a brief overview. For more details, refer to \cite{agrawal2017discrete, ames2019control}.

Consider a set $\set \subset \mathbf{X} \cap \S$ defined as:
\begin{align}
\label{eqn:setI}
    \set := \{\x_k \in \mathbf{X} \cap \S \mid H(\x_k) \geq 0\} \\
    \partial\set := \{ \x_k \in \mathbf{X} \cap \S \mid H(\x_k) = 0\} 
\end{align}
for a smooth function $H: \R^n \to \R$ associated with a Barrier function \cite{ames2014control}. 

\begin{definition}
    (Forward Invariance \cite{ames2019control}). The set $\set \subset \mathbf{X} \cap \S$ is \textit{forward invariant} with respect to \eqref{eq: reduceddiscrete} if:
    \begin{align*}
        \x_0 \in \set ~\Longrightarrow~ \x_k \in \set, ~\forall k \in \N.
    \end{align*}
    The discrete-time hybrid system is \textit{safe} with respect to perturbed initial conditions belonging to the set $\set$  if the set $\set$ is forward invariant with respect to \eqref{eq: reduceddiscrete}.
\end{definition}

While forward invariance is a desirable property for many systems, it can be a challenging property to check in practice. This motivates the use of Barrier functions as a tool for verifying forward invariance. 
Since we have a discrete-time system, we leverage discrete-time barrier functions, originally introduced in \cite{agrawal2017discrete}. Specifically, we use the following definition.

\begin{definition}

    (Discrete-time Barrier Function \cite{ahmadi2019safe}). A function $H: \set \subset \mathbf{X} \cap \S \to \R$ is a \emph{discrete-time barrier function} for the restricted Poincare\'{e} map $\x_{k+1} = P_{\mathbf{X}}(\x_k)$ on the set $\set$ defined by \eqref{eqn:setI} if there exists $\alpha \in (0,1)$ such that for all $\x \in \set$:
    \begin{align}
        H(P_{\mathbf{X}}(\x)) - H(\x) \geq -\alpha H(\x)
    \end{align}
\end{definition}

Note that this condition mimics the form of discrete-time Lyapunov functions, with important differences.  Namely, we do not require $H$ to be positive definite; this is a consequence that we only require set invariance, and not stability.  Yet, because $H$ takes values in the real line (rather than the positive reals), it does imply stability of the set.  This is encoded in the following theorem (which is a straightforward application of the results from \cite{ahmadi2019safe} to the setting of restricted Poincar\'{e} maps). 

\begin{theorem}
\textit{
Let $\x_{k+1} = P_{\mathbf{X}}(\x_k)$ be the Poincar\'{e} map restricted to the set $\mathbf{X} \cap \S$.  If there exists a discrete-time barrier function for the set $\set$, then the set $\set$ is forward invariant and exponentially stable. 
If $\dim(\mathbf{X} \cap \S) = \dim(\S)$ and $\set = \{x^*\}$ then the point $x^*$ is exponentially stable, i.e., the associated periodic orbit is exponentially stable. 
}
\end{theorem}
Practically, in this paper we choose to describe our set $\set$ as a ball of radius $r \in \R^+$, centered around the fixed point of our nominal periodic orbit $\mathcal{O}$ as defined in \eqref{eq: orbit}, but restricted to the reduced-order surface $\mathbf{X}$, i.e.: 
$$
\set := \{\x \mid \x \in \B_{r}(\x^*) \subset \mathbf{X} \cap \S \},
$$
with $\x^* := \Phi(x^*)$. Thus, the statement of hybrid forward invariance of $\set$ is equivalent to the set-based condition  $ P_{\x} ( \B_r(\x^*)) \subseteq \B_r(\x^*)$. Using this set definition, we can also explicitly construct our discrete-time Barrier function as:
\begin{align}
    H(\x) := r - \|\x - (\x^*)\|^2.
\end{align}

\newsec{Simulation-Based Sampling}
To identify the largest set $\set$ that is forward invariant for a given generated nominal limit cycle, we use a sampling-based approach to solve for the largest set $\set$ satisfying the forward-invariance condition. This is framed as an optimization problem of the form:
\begin{align}
    r^* = \max_{r \in [0,r_{\max}]} ~ &r \label{opt: ropt}\\
    \textrm{s.t.} \quad &H(P_{\mathbf{X}}(\x)) - H(\x) \geq -\alpha H(\x), \notag \\
    &\forall \x \in \B_r(\x^*) \notag
\end{align}

While we were proposing the use of $r_\text{opt}$ as a metric of robustness since it characterizes the set that is maximally forward-invariant, it is important to note that the parameter $\alpha$ is also indicative of robustness. In essence, the parameter $\alpha$ indicates how fast the step-to-step dynamics are allowed to approach the boundary of the forward-invariant set.
When designing a control barrier function, the value of $\alpha$ is used as a parameter to control the rate at which the system is allowed to approach the boundary of the safe set, with the system becoming more conservative as $\alpha \to 1$. Since we are concerned with estimating the forward invariant set rather than controlling for safety, we set $\alpha$ close to 0.

\newsec{Simulation-in-the-loop Optimization}
Solving the aforementioned optimization problem not only provides us with the set $\set$ that is hybrid forward invariant, but it also provides us with a measure of robustness for the associated gait. 
It is important to note that this application of discrete-time barrier functions to measure locomotive robustness is independent of the choice of controller. For example, this metric could be used in a RL setting as part of the reward composition to optimize the learned policies for robustness. 


\section{Results}
\label{sec: results}
To show the benefits of the proposed metric, we specifically demonstrate its use towards optimizing offline gaits using a simulation-in-the-loop based framework. This procedure, as illustrated in Fig. \ref{fig: diagram}, is repeated for both flat-foot and multi-contact walking on the Atalante lower-body exoskeleton. 
To further elucidate the efficacy of our approach, we also compare our results to gaits optimized only for stability. A video of the experimental results can be found at \cite{video}. 

\begin{figure}
    \centering
    \includegraphics[width=0.95\linewidth]{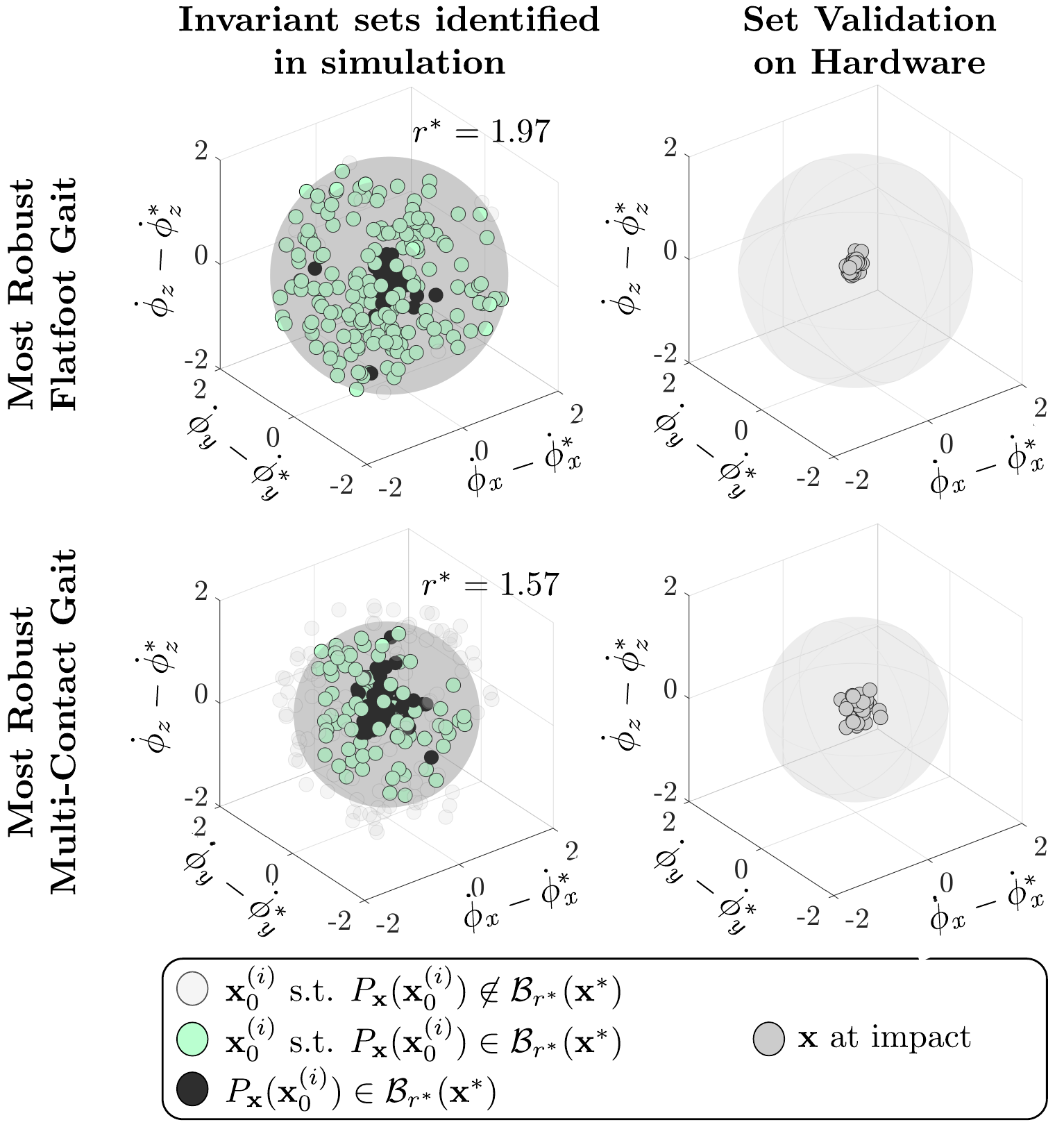}
    \caption{Invariant sets identified in simulation compared to the values seen on hardware during the experiments.}
    \label{fig: sets}
    \vspace{-2mm}
\end{figure}

\begin{figure}
    \centering
    \includegraphics[width=0.9\linewidth]{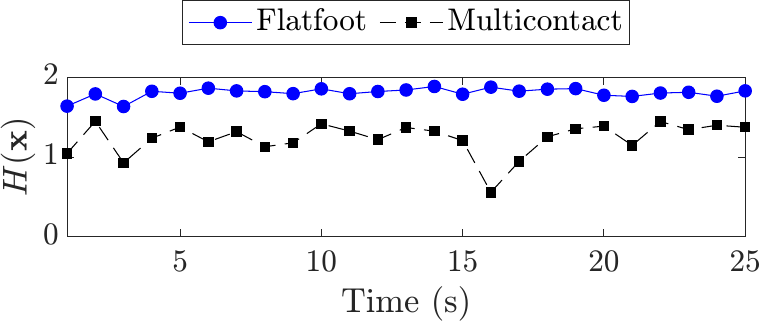}
    \caption{Barrier function evaluation using experimental data}
    \label{fig: exp_barrier_eval}
    \vspace{-2mm}
\end{figure}

\begin{figure*}
    \centering
    \includegraphics[width=\linewidth]{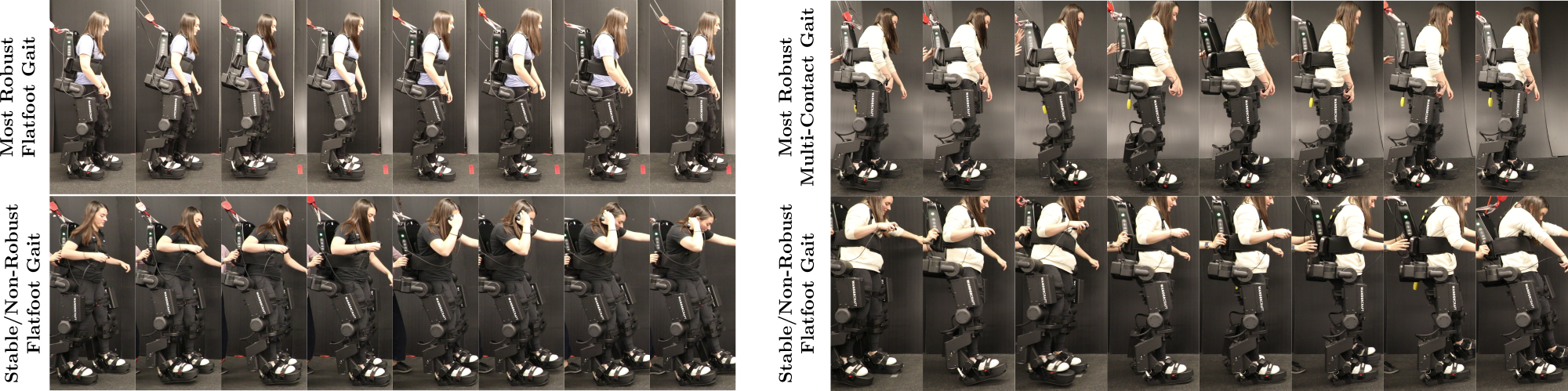}    
    \caption{Experimental Gait Tiles}
    \label{fig: gaittiles}
    \vspace{-2mm}
\end{figure*}

\subsection{Implementation Details}
We choose to generate gaits using the open-source toolbox FROST \cite{hereid2017frost}. These gaits are parameterized using a set of \textit{essential constraints}, as explained in \cite{tucker2021preference}. Specifically, we define these essential constraints to enforce four gait features: step length, step duration, step width, and step height. The output of the gait generation problem is selected to be $7^{th}$-order B\'ezier polynomials, with the phasing variable parameterized by time, for each of the 12 joints of the Atalante lower-body exoskeleton, i.e., $\beta^* \in \R^{12 \times 8}$. These gaits are enforced on the robot by tracking the generated output trajectories. 


Once generated, the gait is then provided to a MuJoCo simulation environment \cite{todorov2012mujoco}. This simulation environment is first used to evaluate whether the simulated locomotion is periodically stable. If the resulting locomotion is stable (empirically evaluated as the total number of steps taken in simulation), this provides us with a fixed point, i.e. $\x^* = P_{\mathbf{X}}(\x^*)$, around which to conduct perturbation analysis. Specifically, a collection of $N = 200$ samples are uniformly drawn from $\B_{r_{\max}}(\x^*)$, with $r_{\max} = 2$. This set is denoted:
$$C_0 := \{\x_0^{(i)} \sim U(\B_{r_{\max}}(\x^*))  \mid i = [1,N] \}.$$ 
To obtain the Poincar\'e return map of the samples (i.e., $ C_P := \{P_{\mathbf{X}}(\x) \mid \forall \x \in C_0\}$ ), we define our reconstruction map $\iota(\x)$ by enforcing the systems outputs and holonomic constraints. Each initial condition is then simulated forward for one gait cycle to obtain the Poincar\'e return set $C_P$. The maximum set that is forward invariant can then be solved for using the optimization problem \eqref{opt: ropt}.

The robustness measure $r^*$, along with the parameters associated with the generated gait, is then provided to a learning algorithm. Here, we choose to leverage preference-based learning since it is able to balance feedback regarding both stability (evaluated by number of steps taken) with robustness (evaluated by $r^*$) without explicit term weighting. The preference feedback is automatically determined to prefer gaits with higher values of $r^*$. For gaits that are not stable, the algorithm automatically prefers gaits with higher number of steps taken. The POLAR toolbox was used to implement the preference-based learning algorithm \cite{tucker2022polar}.

The entire simulation-in-the-loop procedure was conducted for 10 iterations, with 5 gaits being generated and ranked in each iteration. At the conclusion of the framework, the gait identified as being \textit{maximally robust} was then experimentally demonstrated on the Atalante exoskeleton. The same c++ controller used in simulation is used in the experiments, with additional code to interface with Wandercraft API. For the flat-foot behavior a passivity-based controller \cite{gong2022zero} was implemented to track the generated gaits. For the multi-contact behavior, due to the uncertain domain transition resulted changing holonomic constraints, we switch the controller to a joint-level PD controller. The controllers are run directly on the Atalante on-board computer (i5-4300U CPU @1.90GHz with 8GB RAM at 1kHz).





\subsection{Flat-Foot Walking}
The framework was first conducted for flat-foot walking. This domain structure, as illustrated in Fig. \ref{fig: flatfoot-domain}, only has a single domain and a single edge. The domain characterizes the continuous-time dynamics associated with single-support flat-foot walking, with the associated edge characterized by the impact of the non-stance foot. Specifically, the domain is described by the following holonomic constraints:
\begin{align}
    \eta_{SS}(q_e) := \begin{bmatrix}
        p_{st}(q_e) \\
        \phi_{st}(q_e) \label{eq: J_SS}
    \end{bmatrix},
\end{align}
with $p_{st}(q_e) \in \R^3$ and $\phi_{st}(q_e) \in SO(3)$ denoting the position and orientation of the stance foot relative to the world frame. This holonomic constraint is imposed in the gait generation framework via the condition $\eta_{SS}(q_e) = \textrm{constant}$. We refer the reader to \cite{grizzle2014models} for more details.

The entire learning procedure detailed in the previous subsection, was conducted to generate a total of 26 unique gaits. The invariant set associated with the gait being identified as most robust ($r^* = 1.97$ with $r_{\max} = 2$) is illustrated in Fig. \ref{fig: sets}. Additionally, the discrete-time barrier function, evaluated using experimental data, is illustrated in Fig. \ref{fig: exp_barrier_eval}. 

\subsection{Multi-Contact Walking}
To further demonstrate the proposed framework, we repeated the similar process for multi-contact walking. Multi-contact refers to behaviors that can be characterized by changing contact modes throughout a single stride. In the HZD framework, each domain is defined by a unique set of holonomic constraints and a corresponding impact event. While multi-contact walking can include as many 8 unique domains \cite{li2022natural}, we simplify our multi-contact domain structure to only have 2 domains, as illustrated in Fig. \ref{fig: multi-domain}. The domains include the single-support phase (as in flat-foot walking) captured by the holonomic constraint \eqref{eq: J_SS}, as well as an additional domain for the foot-rolling phase captured by the following holonomic constraint:
\begin{align}
    \eta_{DS}(q_e) := [p_{t}(q_e)^{\top}, 
        \phi^x_{t}(q_e), 
        \phi^z_{t}(q_e), \dots \nonumber\\
        p_{h}(q_e)^{\top},
        \phi^x_{h}(q_e),
        \phi^z_{h}(q_e)],  \label{eq: J_DS}
\end{align}
with $p_{t}(q_e) \in \R^3$ denoting the position of the back-foot toe frame with the frame's roll and yaw denoted by $\phi^y_{t}(q_e), \phi^z_{t}(q_e) \in \R$. Similarly, $p_{h}(q_e) \in \R^3$, $\phi^y_{h}(q_e), \phi^z_{h}(q_e) \in \R$ denote the position, roll, and yaw of the  front-foot heel frame. 

Again, the entire learning procedure was conducted for 10 iterations, resulting in a total of 21 unique gaits. The invariant set associated with the most robust gait ($r^* = 1.57$ with $r_{\max} = 2$) is illustrated in Fig. \ref{fig: sets}, with the experimental discrete-time barrier function evaluation shown in Fig. \ref{fig: exp_barrier_eval}.

\subsection{Comparison to Optimizing for Stability}
To further test our proposed metric, we ran an additional set of experiments in which the gaits were optimized for stability. This was done by replacing the metric provided to the learning agent with the total number of steps taken in the simulation environment, rather than $r^*$. These experiments were conducted for both the flat-foot and multi-contact behaviors, with the resulting gaits `optimized for stability' illustrated in Fig. \ref{fig: gaittiles}. It is interesting to note that while these gaits were both stable in simulation, they were not able to successfully yield stable locomotion when translated to hardware. This indicates that our proposed approach is a successful metric for capturing real-world robustness.

\section{Conclusion}
This work explored the use of discrete-time barrier functions to synthesize robust walking gaits. The main idea was that locomotive robustness can be related to forward-invariance of the discrete-time step-to-step dynamics. Specifically, the size of these forward-invariant sets was proposed as a metric for locomotive robustness. Lastly, a simulation-based framework was outlined and demonstrated towards experimentally synthesizing robust nominal gaits for both flat-foot and  multi-contact walking on the Atalante lower-limb exoskeleton. A limitation of this work includes the fact that robustness is not the only factor important for desirable robotic locomotion. As such, future work includes the combination of robustness with other important metrics such as user comfort and naturalness. 

\small{
\section*{ACKNOWLEDGMENTS}
The authors would like to thank the entire Wandercraft team for their continued guidance and technical support with Atalante.}

\bibliographystyle{IEEEtran}
\balance
\bibliography{./Bibliography/IEEEabrv, ./Bibliography/References}


\end{document}